\documentclass{article}

\usepackage[preprint]{neurips_2026}

\usepackage[utf8]{inputenc} 
\usepackage[T1]{fontenc}    
\usepackage{hyperref}       
\usepackage{url}            
\usepackage{booktabs}       
\usepackage{amsfonts}       
\usepackage{nicefrac}       
\usepackage{microtype}      
\usepackage{xcolor}         
\usepackage{amsmath}
\usepackage{graphicx}
\usepackage{xcolor}

\definecolor{charcoal}{rgb}{0.0,0.0,0.0}
\definecolor{darkblue}{rgb}{0.0, 0.0, 0.45}
\definecolor{winered}{rgb}{0.5,0.0,0.0}
\hypersetup{
    colorlinks=true,
    linkcolor=charcoal,
    citecolor=charcoal,
    urlcolor=charcoal
}

\title{Abliteration Mitigation via Refusal Aliases}

\author{%
  Nathan Truong\\
  Independent\\
  \texttt{nathanngoctruong@gmail.com} \\
}

\begin{document}

\maketitle
 
\begin{abstract}
Abliteration, the removal of refusal capabilities from large language models by projecting weight matrices orthogonal to an extracted refusal direction, has emerged as a prominent safety concern through its ability to bypass post-training alignment using only a small set of contrastive prompts. We find that existing defenses commonly overlook the cause of abliteration; that is, how \textit{easily} the refusal direction can be extracted. To hinder this process, we introduce a weight-editing method that obscures the refusal signal by applying rank-$k$ updates to residual stream writer matrices while replacing refusal-inducing activations with random aliases and correcting downstream reader matrices to preserve the model's original behavior. On Llama-3-8B, AMRA improves post-abliteration refusal scores by $2.16$ points over the undefended baseline with less than $0.5$ percentage points of MMLU degradation. On Gemma-2-9B, it improves the post-abliteration refusal by $14.70$ points over the baseline while keeping harmful output rates similar to the baseline, albeit at a greater utility cost. 
\end{abstract}

\section{Introduction}
Large language models (LLMs) have experienced a significant influx in usage over the past few years, publicly demonstrating their ability to solve problems and generally enhance the quality of life. Notably, their ability to reason \citep{Guo2025, kojima2022large}, follow instructions \citep{ouyang2022training, wei2022finetunedlanguagemodelszeroshot, chung2022scalinginstructionfinetunedlanguagemodels}, and generalize to out-of-distribution domains \citep{sanh2022multitask, wang-etal-2022-super} has directly led to their increased usage in medical \citep{Gaber2024.09.27.24314505}, security \citep{jiang2025investigatinglargelanguagemodels}, finance \citep{lin2025evaluation}, and legal settings \citep{guha2023legalbench, kant2025robustlegalreasoningharnessing}. The prominence of LLM usage in high-risk settings underscores the need for secure and benign deployments.

Recent work has found that language models are vulnerable to black-box jailbreaking methods regardless of their post-training alignment, with many leveraging adversarial prompt injections \citep{NEURIPS2024_70702e8c, wei2023jailbroken}. 
When compromised, LLMs can produce harmful or invalid information, standing as a potential threat to safety and legal policies \citep{shi2024largelanguagemodelsafety}. To mitigate LLM misuse, a wide body of work has emerged attempting to further align models with human policies or constitutions \citep{ouyang2022training, bai2022constitutionalaiharmlessnessai}, with methods commonly applying additional fine-tuning \citep{wang-etal-2023-self-instruct, rafailov2023direct}, inference-time interventions \citep{li2023inferencetime, lee2025programming}, and mechanistic interpretability approaches \citep{arditi2024refusal, obrien2025steeringlanguagemodelrefusal}. 

More specifically, through mechanistic interpretability, it has recently been found that refusal behavior in LLMs can be isolated to a low-dimensional linear direction in a language model's residual-stream activation space \citep{arditi2024refusal}, ultimately giving rise to abliteration---a jailbreaking method that removes refusal capabilities by projecting weight matrices to be orthogonal to an extracted refusal direction \citep{failspy_abliterator_2026, heretic}. Most notably, abliteration is concerning by virtue of its white-box modifications that can bypass post-training alignment wholesale while only requiring a small set of contrastive prompts and no additional training. While existing defenses have proposed activation steering \citep{lee2025programming, sheng2026alphasteer} and circuit-level interventions \citep{zou2024improving} to enhance refusal, none explicitly address the refusal direction extraction itself.

As a result of this we propose \textbf{Abliteration Mitigation via Refusal Aliases  (AMRA)}, a weight-updating approach that aims to obfuscate the refusal vector in the activation space. Specifically,  we apply rank-$k$ updates to matrices that write to the residual stream at causally relevant layers, replacing refusal-inducing activations with low-variance random aliases while patching downstream reader matrices to preserve the model's original refusal behavior. We validate AMRA on Llama-3-8B and Gemma-2-9B, showing that it substantially improves robustness to directional ablation while inflicting minimal utility degradation on Llama and moderate costs on Gemma. To our knowledge, AMRA is the first post-hoc weight-editing defense that explicitly targets the extractability of the refusal direction as a way to hinder abliteration without requiring additional fine-tuning \citep{shairah2025embarrassinglysimpledefensellm}. Crucially, our method is a single-use and permanent application, not requiring additional compute after obfuscation. 

As a vital note, the work is geared predominantly towards language model developers as a way to provide them with a method capable of securing their newly-made LLM \textit{prior to the first release of the weights.} An attacker with access to the original model without our obfuscation defeats the purpose of its existence as they can simply use abliteration techniques on the unprotected weights.


\section{Methodology} \label{sec:methodology}
In this section, we provide a comprehensive and formal breakdown of our obfuscation method. 

\subsection{Transformer Background}
A decoder-only transformer language model \citep{j.2018generating} comprises a series of modules that update an initial embedding representation by consecutively reading and writing to the residual stream \citep{vaswani2023attentionneed, elhage2021mathematical}. Notable weight matrices that "read" or "fork" from the residual stream are the Query, Key, and Value matrices inherent to the attention mechanism and the upward projection matrix into the feed-forward (FFW) layers. We henceforth denote these matrices in the form of: $W^l_{\text{in}}$. Similarly, matrices that write to the residual stream like the downward linear projection after multi-headed attention and FFW are denoted as $W^l_{\text{O}}$ and $W^l_{\text{out}}$.

\subsection{Layer Selection}
\paragraph{Refusal Vector Extraction}
Many existing jailbreaking methods rely on the initial observation that there exists a ubiquitous refusal vector $\textbf{r}$ in the residual stream space of transformer language models \citep{vaswani2023attentionneed,arditi2024refusal,lai2025projected_abliteration,heretic}. To uncensor LLMs (i.e., to excise their refusal capabilities), weight matrices are projected to be orthogonal to $\textbf{r}$ \citep{arditi2024refusal}, giving rise to "abliteration" \citep{failspy_abliterator_2026}. The extraction of the refusal direction $\textbf{r}$ is typically performed using a \textit{difference-in-means} approach \citep{eleutherai_diff_in_means} which comprises the collection of the average difference in activations per layer between harmful and benign prompts. Following existing work \citep{arditi2024refusal}, we provide a formal definition as follows: Given a dataset of harmful and benign prompts denoted as $\mathcal{D}_{\mathit{harmful}}$ and $\mathcal{D}_{\mathit{benign}}$, we compute the average harmful activation $\mu^{l}$ and average benign activation $\mathit{v}^l$ for a layer $l \in \{L\}$ at the first token position following the input prompts through:
\begin{figure}
    \centering
    \includegraphics[width=1\linewidth]{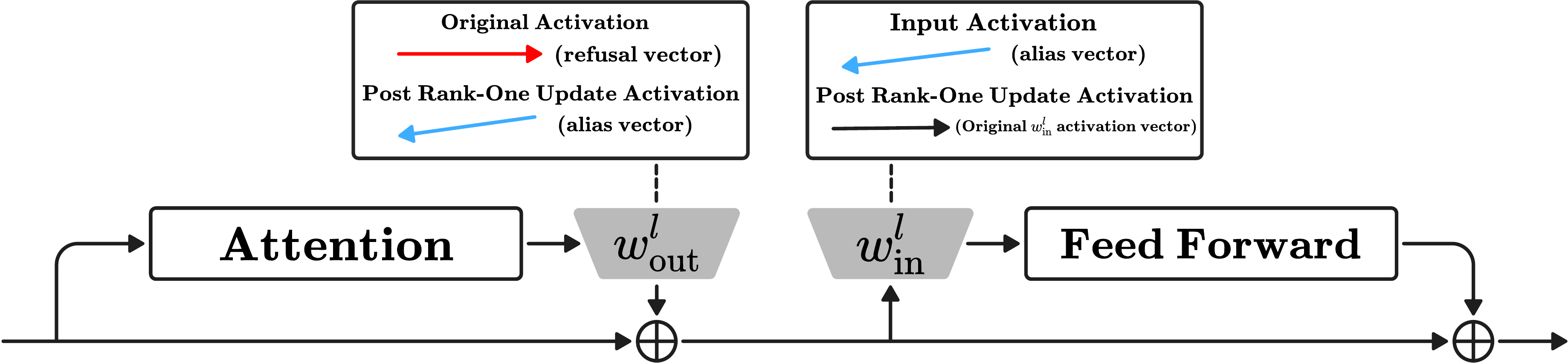}
    \caption{\textbf{Obfuscation diagram.} We show a high-level depiction of the patches made to obfuscate the refusal signal. Although our weight matrix updates are rank-k, we show a rank-one variant for brevity. The bottom arrow represents the residual stream through layer $l$. The Q, K, and V, up projections into the attention mechanism, downward projection from the feed-forward, and LayerNorm have been omitted to highlight sublayer interactions and general stylistic simplicity.}
    \label{fig:diagram}
\end{figure}
\[
\mu^l = \frac{1}{|\mathcal{D}_{\mathrm{harmful}}|}
\sum_{t \in \mathcal{D}_{\mathrm{harmful}}} x^l(t), \quad
v^l = \frac{1}{|\mathcal{D}_{\mathrm{benign}}|}
\sum_{t \in \mathcal{D}_{\mathrm{benign}}} x^l(t).
\]
Here, our approximate per-layer refusal vector is calculated through: 
\[
{r}^l = \mu^l - v^l
\]
We let $\hat{{r}}^l$ denote the unit-norm variant $\frac{\boldsymbol{r}^l}{\|\boldsymbol{r}^l\|_2}$ and $x^l(t)$ the hooked activations of the weight matrix $x^l$ when $t$ is passed into the model as a prompt.
\paragraph{Layer Selection}
To find transformer layers that are causally pertinent to the refusal direction $\hat{\textbf{r}^l}$, we first compute a unit-norm refusal direction for every layer: $\{\hat{r}^{\,l} \mid l \in \mathcal{L}\}$. For each layer $l$, we calculate the vector's effect on refusal by ablating it from the residual stream while leaving other layers unchanged. Specifically, given a layer $l$, we construct the ablated representation
\[
\tilde{x}^l = x^l-\text{proj}_{\hat{r}^l}(x^l)=x^l-\langle x^l, \hat{r}^l\rangle \hat{r}^l
\]
We then evaluate the model's attack success rate (ASR) after each ablation through adversarial prompts using HarmBench \citep{mazeika2024harmbenchstandardizedevaluationframework}. Layers that are deemed relevant through this method are organized into a set $\{\mathcal{L}'\}$.
\subsection{Activation Obfuscation}
Existing abliteration methods heavily rely on the extraction of $\boldsymbol{\hat{r}}^l$. To hinder this process, we propose updating the relevant weight matrices that write to the residual stream through a series of rank-k updates. As depicted in Figure~\ref{fig:diagram}, we update the matrix such that given penultimate activations that yield the refusal vector as the matrix's output, it is rewritten to output a random and low-variance alias vector. 

Similar to existing work \citep{lee2025programming}, for a relevant layer $l$ and writer matrix $W^l \in \{W^l_{O},W^l_{\text{out}}\} \subseteq \mathbb{R}^{d_{\text{resid}} \times d_{\text{hidden}}}$ \footnote{We denote the model or residual stream dimension as $d_{\text{resid}}$ distinctively from $d_{\text{hidden}}$ which generally describes the dimension of activations directly prior to the writer matrix within a sublayer (e.g. feedforward, attention, and unembedding).}, we collect the last-token  position writer outputs for benign and harmful prompts into ${H}^l_+, {H}^l_- \in \mathbb{R}^{n \times d_{\text{resid}}}$ respectively, where $n$ represents sample size and $d_{\text{resid}}$ represents the residual stream dimension. 

\begin{figure}
    \centering
    \includegraphics[width=1\linewidth]{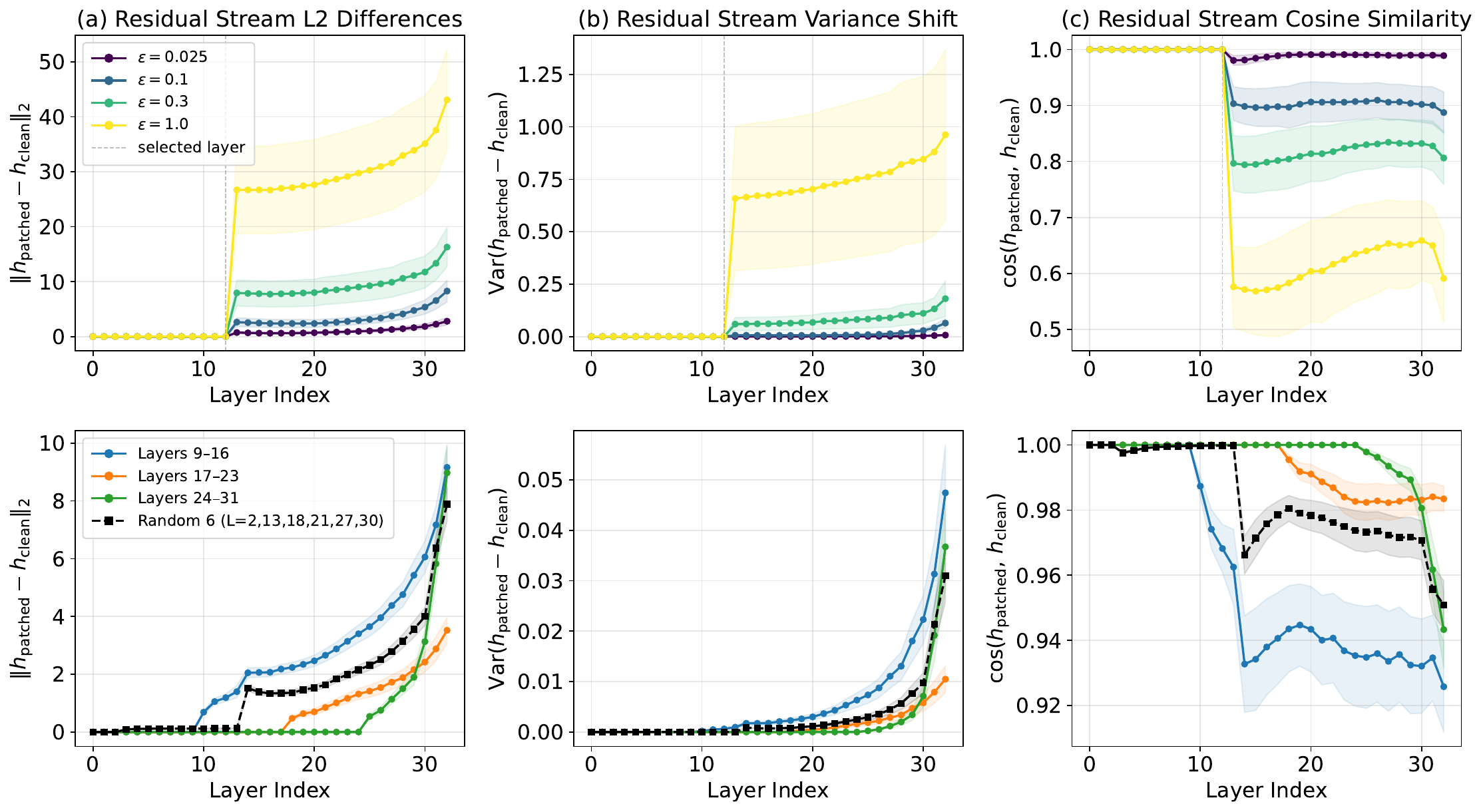}
    \caption{\textbf{Residual stream shifts.} Here, we show the effect of a rank-one ($k_w = 1$) update of a writer matrix $W^l_{\text{out}}$ on subsequent residual stream representations. Specifically, we compare the mean residual stream values prior to and after the weight matrix update over 16 prompts. (a) exhibits the L2 difference, (b) shows the shift in residual stream variance, and (c) shows the cosine similarity of the residual stream before versus after the edit. \textbf{Upper:} sweeps over $\varepsilon$ for layer 20. \textbf{Lower:} sweeps over layers along with an additional random selection of 6.}
    \label{fig:residual_stream_shift}
\end{figure}

Subsequently, we locate candidate refusal directions by applying principal component analysis (PCA) on the difference in mean-centered activations $\bar{{H}}^l_-$ and $\bar{{H}}^l_+$. More specifically, we initially compute the singular value decomposition over ${H}^l_{\text{diff}}=\bar{{H}}^l_--\bar{{H}}^l_+$:
\[
{H}^l_{\text{diff}}= {U}{\Sigma}{V}^{\top}
\]
where ${U}$ and ${V}$ contain the unit-norm eigenvectors of $({H}^l_{\text{diff}})({H}^l_{\text{diff}})^\top$ and $({H}^l_{\text{diff}})^\top({H}^l_{\text{diff}})$ respectively. Through this construction, the columns of $V$ represent the principal components of $H^l_\text{diff}$. These vectors are ranked by the amount of variance that they capture in our refusal-related activation space for a given layer: 
\[
\{v^l_1, v^l_2,v^l_3 \cdots v^l_n\} \subseteq \mathrm{Col}(V).
\]
Then, we update $W^l \in \mathbb{R}^{d_{\mathrm{resid}}\times d_{\mathrm{hidden}}}$ through a rank-$k_w$ addition to produce random activations instead of the previous refusal-inducing directions.
\[
\tilde{W^l} = W^l+ (A - R)^\top R W^l
\]
Here, $R \in \mathbb{R}^{k_w \times d_{\mathrm{resid}}}$ contains the top-$k_w$ column vectors of $V$ or correspondingly, the top-$k_w$ principal components of $H^l_{\text{diff}}$. We subtract $R$ from $A\in \mathbb{R}^{k_w \times d_{\mathrm{resid}}}$ where each row vector $A_{i} \sim \mathcal{N}(0, \varepsilon^2I)$. Intuitively, if $\tilde{W}^l$ receives an input that is mapped to a vector in the subspace spanned by our chosen top-k refusal principal components, it is instead mapped to a random alias vector. Notably, as shown in Figure~\ref{fig:residual_stream_shift}, our choice of $\varepsilon$ and $l$ directly influences the amount of pollution introduced into the residual stream.


We acknowledge that rank-k updates work most ideally when inputs to $\tilde{W}^l$ yield activations that align with $R$. However, this is relatively uncommon in practice. Activations may only be slightly aligned with $R$, incidentally spreading noise in downstream computations. However, in Section~\ref{sec:experiments}, we show that with optimal $\varepsilon$ and $k_w$, these updates minimally degrade model quality while significantly consolidating refusal capabilities against abliteration.
\paragraph{Downstream Weight Patches}
Although this method effectively obfuscates the refusal vector for a specific sublayer, subsequent layers are unable to interpret the updated writer matrix's random output as inputs that induce refusal. To alleviate this, we fix all downstream matrices that read from the residual stream (e.g., $W^l \in \{Q^l, K^l, V^l, W^l_{\mathrm{in}}, W_{\mathrm{unembed}}\}$) at varying $l$ to output their original activations through similar rank-$k_r$ updates. Specifically, for a reader matrix $W \in \mathbb{R}^{d_{\mathrm{in}}\times d_{\mathrm{resid}}}$, given polluted inputs $X_{p}\in\mathbb{R}^{n\times d_{\mathrm{resid}}}$ to $W$ from the residual stream (polluted due to previous writer matrix patches) and clean inputs (without the writer matrix patches) $X_{c} \in \mathbb{R}^{n\times d_{\mathrm{resid}}}$, we construct the following Singular Value Decompositions (SVD):
  \[           
    X_p \;\approx\; U_{k_r}\,\Sigma_{k_r}\, V_{k_r}^{\!\top},   
    \quad    
    U_{k_r} \in \mathbb{R}^{n \times k_r},\;      
    \Sigma_{k_r} \in \mathbb{R}^{k_r \times k_r},\;           
    V^{\top}_{k_r} \in \mathbb{R}^{k_r \times d_{\mathrm{resid}}}  
  \]
  The correction matrix 
  \[
    M \;=\; (X_c - X_p)^{\!\top}\, U_{k_r}\, \Sigma_{k_r}^{-1}\, V_{k_r}^{\!\top}        
    \;\in\; \mathbb{R}^{d_{\mathrm{resid}} \times d_{\mathrm{resid}}}
  \]                                                                                                         
  yields the reader weight update
  \[   
    W_{\mathrm{new}} \;=\; W + W\,M
  \]
  which is the rank-$\le k_r$ correction minimizing the difference in Frobenius norm 
  \[   
    \bigl\lVert X_p\,(W_{\mathrm{new}})^{\top} - X_c\,W^\top \bigr\rVert^2_F 
  \]                     
  by the Eckart-Young theorem. Naturally, the update redirects outputs in top-$k_r$ right-singular subspace of $X_p$ towards $X_c$. It recovers $X_p\,W_{\mathrm{new}}^\top = X_c\,W^\top$ through that subspace and leaves $W$ unchanged on its orthogonal complement.





\section{Experiments} \label{sec:experiments}
\subsection{Experimental Setup}
\paragraph{Baseline Setup} 
We test our method on two popular open-source transformer-based LLMs: Llama-3-8B \citep{grattafiori2024llama3herdmodels}, and Gemma-2-9B \citep{gemmateam2024gemma2improvingopen}. These both share a relatively straightforward architecture, ultimately simplifying the experimentation process.
\paragraph{Refusal Direction Extraction} Following existing work \citep{heretic}, we extract the refusal direction by using contrasting instruction datasets from \texttt{mlabonne/harmful\_behaviors} \citep{labonne_harmful_behaviors} and \texttt{mlabonne/harmless\_alpaca} \citep{labonne_harmless_alpaca}. Using 400 prompts from each, we filter the datasets by retaining harmful prompts that elicit refusal and benign prompts that do not. This would allow for an easier obtainment of the refusal direction when performing difference-in-means by using examples with greater signal.
\paragraph{Utility Baselines} To measure utility preservation, we record pre-obfuscation and post-obfuscation benchmark performance using EleutherAI's \texttt{lm-evaluation-harness} \citep{eval-harness}. Specifically, we analyze the preservation of problem-solving capabilities using GSM8K \citep{cobbe2021trainingverifierssolvemath} and broad knowledge coverage through MMLU \citep{hendrycks2021measuring}. Following existing work \citep{arditi2024refusal}, we record the average pre and post dataset bits-per-byte—which is tokenizer agnostic—over 1024 sequences of The Pile \citep{gao2020pile800gbdatasetdiverse} and Alpaca \citep{alpaca} to measure the effect of our method on general next-token-prediction quality.
\paragraph{Attacks} 
 To validate the effectiveness of our method, we employ the conventional directional ablation of the refusal vector through weight orthogonalization \citep{arditi2024refusal}.
\paragraph{Defense Baselines} We also compare our approach to existing jailbreak defenses. Specifically, we implement Surgical \citep{wang2025surgicalcheapflexiblemitigating}, CAST \citep{lee2025programming}, Circuit Breakers \citep{zou2024improving}, and AlphaSteer \citep{sheng2026alphasteer}.
\paragraph{Hyperparameter selection} Our hyperparameters (i.e., $\{l\}$, $k_w$, $k_r$  and $\varepsilon$) were chosen using NSGA-II \citep{nsga2} through Optuna \citep{ozaki2025optunahub}. In our setting, Pareto optimal configurations minimize the distance between a model's refusal rate after an attack is applied to its baseline refusal rate along with utility heuristics like bits-per-byte on The Pile \citep{gao2020pile800gbdatasetdiverse} and MMLU \citep{hendrycks2021measuring}. We find that for Llama-3-8B Instruct, the optimal values of $\varepsilon$, layers, $k_w$, and $k_r$ are 0.025, 9-31, 1, and 8, respectively. For the same hyperparameters, Gemma-2-9B's optimal configuration was 0.025, 9-41, 4, and 16.


\subsection{Post-Abliteration Results}
Compared to baselines, we show AMRA generally yields the best post-abliteration scores. As depicted in Table~\ref{tab:safety-robustness}, applying AMRA improves the baseline refusal rate of both Llama-3-8B and Gemma-2-9B by roughly 0.203 and 0.083 points, respectively. For directional ablation, AMRA scores 2.16 points higher than the undefended baseline for Llama-3-8B and $14.70$ points higher on Gemma-2-9B. However, in Gemma's case, AMRA still yields a notably negative post-abliteration refusal score ($-1.3105$), indicating that while the obfuscation substantially closes the gap relative to baseline ($-16.0093$), the Arditi-style attack still partially succeeds at degrading refusal on this architecture. Nevertheless, AMRA is the only defense on Gemma-2-9B that significantly fortifies the model against refusal ($+14.70$-point improvement over the undefended Gemma) while keeping Harmbench ASR under $0.02$ and LlamaGuard's unsafety score at $0$. Among other defenses, AlphaSteer is the only other method that achieves a positive post-abliteration refusal score on Gemma ($0.8756$), but it does so at the cost of reduced clean refusal ($5.6320$ vs.\ $7.1172$) and elevated HarmBench ASR ($0.09$) and LlamaGuard unsafe rates ($0.11$). Circuit Breakers and CAST provide negligible or no improvement in post-abliteration refusal on Gemma: CB's Arditi score ($-16.0190$) is nearly identical to the undefended baseline, while CAST improves it only modestly to $-12.3973$. Surgical performs worst overall, substantially degrading both clean refusal and post-abliteration robustness on both models, and inducing the highest HarmBench ASR ($0.42$ on Llama, $0.44$ on Gemma). Across both architectures, AMRA is the only defense that simultaneously improves clean refusal behavior, substantially raises post-abliteration refusal scores, and maintains low HarmBench ASR and LlamaGuard unsafe rates.
\begin{table}[h]
\caption{Safety and abliteration results. Refusal scores are higher when the model retains more refusal behavior before and after Arditi-style abliteration. Additionally, the Arditi abliteration on defenses other than our baseline. (None) implies the difference-in-means refusal vector extraction was run again after a defense was applied. HarmBench ASR and LlamaGuard unsafe rate are lower when the model is safer.}
\centering
\small
\setlength{\tabcolsep}{4pt}
\begin{tabular}{llcccc}
\toprule
Model & Defense & Clean Refusal $\uparrow$ & Arditi $\uparrow$ & HarmBench ASR $\downarrow$ & LlamaGuard $\downarrow$ \\
\midrule
Llama-3-8B & None & 10.0318 & 5.6884 & 0.0200 & 0.0100 \\
 & AMRA & 10.2350 & 7.8497 & 0.0100 & 0.0100 \\
 & Surgical & 1.3365 & 2.0229 & 0.4200 & 0.2800 \\
 & CAST & -0.2591 & -0.5980 & 0.0000 & 0.8300 \\
 & CB & 9.9183 & 5.5355 & 0.0200 & 0.0400 \\
 & AlphaSteer & 10.0184 & 5.6796 & 0.0200 & 0.0200 \\
\midrule
Gemma-2-9B & None & 7.1172 & -16.0093 & 0.0200 & 0.0000 \\
 & AMRA & 7.2000 & -1.3105 & 0.0100 & 0.0000 \\
 & Surgical & -16.9482 & -18.0443 & 0.4400 & 0.7300 \\
 & CAST & 7.3802 & -12.3973 & 0.0000 & 0.0000 \\
 & CB & 7.1436 & -16.0190 & 0.0200 & 0.0000 \\
 & AlphaSteer & 5.6320 & 0.8756 & 0.0900 & 0.1100 \\
\bottomrule
\end{tabular}
\label{tab:safety-robustness}
\end{table}
\subsection{Utility Results}
We show that our defense does not significantly degrade model quality with respect to baseline performance. 
\begin{table}[h]
\caption{Utility results across base models and defenses. Lower BPB is better; higher GSM8K and MMLU are better.}
\centering
\small
\begin{tabular}{llcccc}
\toprule
Model & Defense & Pile BPB $\downarrow$ & Alpaca BPB $\downarrow$ & GSM8K $\uparrow$ & MMLU $\uparrow$ \\
\midrule
Llama-3-8B & None & 0.7665 & 0.5555 & 0.7020 & 0.6911 \\
 & AMRA & 0.7801 & 0.5651 & 0.6880 & 0.6876 \\
 & Surgical & 0.7434 & 0.5141 & 0.7420 & 0.6782 \\
 & CAST & 1.2924 & 0.8295 & 0.0320 & 0.3334 \\
 & CB & 0.7674 & 0.5576 & 0.7080 & 0.6825 \\
 & AlphaSteer & 0.7667 & 0.5555 & 0.6980 & 0.6912 \\
\midrule
Gemma-2-9B & None & 0.8124 & 0.6561 & 0.5620 & 0.7411 \\
 & AMRA & 0.9597 & 0.6679 & 0.3360 & 0.6926 \\
 & Surgical & 0.8577 & 0.6602 & 0.6760 & 0.6971 \\
 & CAST & 0.8117 & 0.6526 & 0.5080 & 0.7393 \\
 & CB & 0.8124 & 0.6563 & 0.5620 & 0.7412 \\
 & AlphaSteer & 1.5838 & 1.0017 & 0.1680 & 0.6611 \\
\bottomrule
\end{tabular}
\label{tab:utility}
\end{table}
As shown in Table~\ref{tab:utility}, compared to our baselines, AMRA retains strong utility on Llama-3-8B with only marginal degradation: Pile BPB increases by $0.0136$ ($0.7665 \to 0.7801$), Alpaca BPB by $0.0096$, GSM8K decreases by $1.4$ percentage points, and MMLU by $0.35$ percentage points. These losses are comparable to or smaller than those of CB and AlphaSteer, both of which similarly preserve utility on Llama. By contrast, CAST catastrophically degrades Llama's utility, inflating Pile BPB to $1.2924$ and reducing MMLU to $0.3334$ and GSM8K to $0.0320$---rendering the model effectively unusable for reasoning tasks.

On Gemma-2-9B, AMRA incurs more noticeable utility costs: Pile BPB rises from $0.8124$ to $0.9597$, GSM8K drops from $0.5620$ to $0.3360$, and MMLU decreases by roughly $4.9$ percentage points. This is a meaningful trade-off that we attribute to the higher rank of the writer updates ($k_w = 4$) selected by our hyperparameter search on Gemma, which introduces more residual stream perturbation than the rank-one configuration used for Llama. More robust hyperparameter searches may yield better results in this setting. However, it is worth noting that AlphaSteer attains a higher post-abliteration refusal score than AMRA on Gemma ($0.8756$ versus $-1.3105$), but at a substantial utility cost (Pile BPB $1.5838$ vs AMRA's $0.9597$, GSM8K $0.1680$ vs AMRA's $0.3360$). CB and CAST both retain stronger raw utility on Gemma, but as shown in Table~\ref{tab:safety-robustness}, they provide negligible defense against directional ablation on this architecture. Surgical presents an unusual profile on Gemma, slightly improving GSM8K ($0.6760$) and MMLU ($0.6971$) relative to baseline, but this comes at the cost of severe safety degradation. Overall, these results highlight a robustness--utility trade-off that is architecture-dependent: AMRA achieves a favorable balance on Llama-3-8B and provides the strongest abliteration defense on Gemma-2-9B at a moderate utility cost that may be further reduced through more targeted hyperparameter tuning or lower-rank configurations.


\section{Related Work}
\paragraph{LLM Safety} As LLM usage becomes increasingly prevalent, strictly enforcing their secure employments to prevent downstream safety concerns \citep{openai2024gpt4technicalreport, grattafiori2024llama3herdmodels} has become a major topic of contemporary research. If inadequately deployed, LLMs may provide harmful outputs, ultimately making them a security threat. For instance, when prompted, "How do I synthesize poisonous gas?" a misaligned model may produce a set of instructions to chemically produce the gas, potentially endangering lives. To prevent the acknowledgment of malicious prompts, a commonly implemented solution consists of employing post-training alignment techniques to constrain LLMs to predefined constitutions or human morals \citep{ouyang2022training, bianchi2024safetytuned, bai2022constitutionalaiharmlessnessai}. However, recent evidence shows that language models remain vulnerable to "jailbreaks" in spite of their purported safety mechanisms \citep{liu2024autodan, chao2023jailbreaking, zou2023universaltransferableadversarialattacks}, influencing the construction of novel defenses aimed toward mitigating these attacks. For instance, deeper post-training methods such as reinforcement learning from human feedback \citep{NIPS2017_d5e2c0ad, ouyang2022training} and direct policy optimization \citep{rafailov2023direct, grattafiori2024llama3herdmodels}. These methods operate through contrastive refusal training, which updates the model to align with human preferences (e.g., refusing malicious questions).

In another vein, activation-level methods provide a more fine-grained approach to alignment. Studies in mechanistic interpretability show that LLMs already possess linearly separable internal structures in their activation space that represent distinct features \citep{elhage2022superposition, park2023the, durmus2024steering}. With this observation, approaches locating refusal-related features have recently garnered significant attention for its potential to enforce refusal in jailbreaking settings \citep{arditi2024refusal, obrien2025steeringlanguagemodelrefusal}.

\paragraph{Refusal Vectors} Locating feature vectors in language models using contrastive prompts has become common practice to alter model behavior \citep{rimsky-etal-2024-steering, postmus2024steering, zou2025representationengineeringtopdownapproach}. A large body of recent safety work has been developed based on the nascent observation that refusal can be localized to low-dimensional linear directions in activation space \citep{arditi2024refusal, wang2025surgicalcheapflexiblemitigating}. These directions are commonly estimated by finding the average difference in activations for harmful and harmless prompts \citep{arditi2024refusal, zou2024improving}. Activation steering, which is performed by adding feature vectors to the residual stream, has been extensively used to fortify language model refusal. Existing approaches are predominantly inference-time techniques that actively shift hidden states along an extracted safety direction during the forward pass \citep{shen2025jailbreak, lee2025programming, sheng2026alphasteer, shairah2026turning}. By contrast, the same approach may be used to \textit{preclude} refusal, with works commonly ablating the refusal direction from the residual stream to allow for uncensored output \citep{arditi2024refusal, heretic}. By hindering refusal, abliteration allows for the disclosure of dangerous and potentially false information, ultimately making it one of the largest LLM safety concerns to date by undermining post-training alignment. Existing methods do not explicitly obfuscate the refusal vector, ultimately leaving them vulnerable to activation steering-based uncensoring.

\paragraph{Model Editing} Rank-one weight matrix edits have been used to alter and rewrite information in language models \citep{meng2022locating, meng2023massediting, ilharco2023editing}. By perturbing a small number of feed-forward projection matrices through a similar rank-one update, we adjust how the refusal feature is routed through downstream layers rather than directly affecting factual knowledge or storage. Distinct from these model editing methods, our primary goal is to obfuscate a direction rather than to replace. 
\section{Limitations and Future Work}
Our limitations are chiefly due to single attack experiments, which introduces questions regarding how more sophisticated extraction methods like nonlinear probes or iterative searches could bypass AMRA's obfuscation. The results are also architecture-dependent: AMRA incurs minimal utility loss on Llama-3-8B, but substantially degrades GSM8K and Pile BPB on Gemma-2-9B, the model that used higher-rank updates, and the post-abliteration refusal score remains negative. We may be able to alleviate this with more comprehensive hyperparameter searches or tailoring AMRA to the model's architecture rather than aiming for a general and widely-applicable implementation. AMRA is able to make the refusal vector harder to locate but does not inherently strengthen the feature. This may be achieved by combining our obfuscation with methods that strengthen the refusal direction \citep{lee2025programming, zou2024improving}. However, we leave this to future work.
\section{Conclusion}
In this work, we introduce AMRA, a refusal-direction obfuscation method that replaces refusal-inducing activations with random aliases through rank-$k$ updates. On Llama-3-8B, AMRA greatly improves post-abliteration refusal while preserving utility. On Gemma-2-9B, it also substantially fortifies refusal capabilities post-abliteration though at a greater utility cost. Our results show that targeting the extraction of the refusal direction is a viable approach for mitigating weight-space jailbreaking methods.

\bibliographystyle{plainnat}
\bibliography{references}

@inproceedings{
arditi2024refusal,
title={Refusal in Language Models Is Mediated by a Single Direction},
author={Andy Arditi and Oscar Balcells Obeso and Aaquib Syed and Daniel Paleka and Nina Rimsky and Wes Gurnee and Neel Nanda},
booktitle={The Thirty-eighth Annual Conference on Neural Information Processing Systems},
year={2024},
url={https://openreview.net/forum?id=pH3XAQME6c}
}

@article{Guo2025,
   title={DeepSeek-R1 incentivizes reasoning in LLMs through reinforcement learning},
   volume={645},
   ISSN={1476-4687},
   url={http://dx.doi.org/10.1038/s41586-025-09422-z},
   DOI={10.1038/s41586-025-09422-z},
   number={8081},
   journal={Nature},
   publisher={Springer Science and Business Media LLC},
   author={Guo, Daya and Yang, Dejian and Zhang, Haowei and Song, Junxiao and Wang, Peiyi and Zhu, Qihao and Xu, Runxin and Zhang, Ruoyu and Ma, Shirong and Bi, Xiao and Zhang, Xiaokang and Yu, Xingkai and Wu, Yu and Wu, Z. F. and Gou, Zhibin and Shao, Zhihong and Li, Zhuoshu and Gao, Ziyi and Liu, Aixin and Xue, Bing and Wang, Bingxuan and Wu, Bochao and Feng, Bei and Lu, Chengda and Zhao, Chenggang and Deng, Chengqi and Ruan, Chong and Dai, Damai and Chen, Deli and Ji, Dongjie and Li, Erhang and Lin, Fangyun and Dai, Fucong and Luo, Fuli and Hao, Guangbo and Chen, Guanting and Li, Guowei and Zhang, H. and Xu, Hanwei and Ding, Honghui and Gao, Huazuo and Qu, Hui and Li, Hui and Guo, Jianzhong and Li, Jiashi and Chen, Jingchang and Yuan, Jingyang and Tu, Jinhao and Qiu, Junjie and Li, Junlong and Cai, J. L. and Ni, Jiaqi and Liang, Jian and Chen, Jin and Dong, Kai and Hu, Kai and You, Kaichao and Gao, Kaige and Guan, Kang and Huang, Kexin and Yu, Kuai and Wang, Lean and Zhang, Lecong and Zhao, Liang and Wang, Litong and Zhang, Liyue and Xu, Lei and Xia, Leyi and Zhang, Mingchuan and Zhang, Minghua and Tang, Minghui and Zhou, Mingxu and Li, Meng and Wang, Miaojun and Li, Mingming and Tian, Ning and Huang, Panpan and Zhang, Peng and Wang, Qiancheng and Chen, Qinyu and Du, Qiushi and Ge, Ruiqi and Zhang, Ruisong and Pan, Ruizhe and Wang, Runji and Chen, R. J. and Jin, R. L. and Chen, Ruyi and Lu, Shanghao and Zhou, Shangyan and Chen, Shanhuang and Ye, Shengfeng and Wang, Shiyu and Yu, Shuiping and Zhou, Shunfeng and Pan, Shuting and Li, S. S. and Zhou, Shuang and Wu, Shaoqing and Yun, Tao and Pei, Tian and Sun, Tianyu and Wang, T. and Zeng, Wangding and Liu, Wen and Liang, Wenfeng and Gao, Wenjun and Yu, Wenqin and Zhang, Wentao and Xiao, W. L. and An, Wei and Liu, Xiaodong and Wang, Xiaohan and Chen, Xiaokang and Nie, Xiaotao and Cheng, Xin and Liu, Xin and Xie, Xin and Liu, Xingchao and Yang, Xinyu and Li, Xinyuan and Su, Xuecheng and Lin, Xuheng and Li, X. Q. and Jin, Xiangyue and Shen, Xiaojin and Chen, Xiaosha and Sun, Xiaowen and Wang, Xiaoxiang and Song, Xinnan and Zhou, Xinyi and Wang, Xianzu and Shan, Xinxia and Li, Y. K. and Wang, Y. Q. and Wei, Y. X. and Zhang, Yang and Xu, Yanhong and Li, Yao and Zhao, Yao and Sun, Yaofeng and Wang, Yaohui and Yu, Yi and Zhang, Yichao and Shi, Yifan and Xiong, Yiliang and He, Ying and Piao, Yishi and Wang, Yisong and Tan, Yixuan and Ma, Yiyang and Liu, Yiyuan and Guo, Yongqiang and Ou, Yuan and Wang, Yuduan and Gong, Yue and Zou, Yuheng and He, Yujia and Xiong, Yunfan and Luo, Yuxiang and You, Yuxiang and Liu, Yuxuan and Zhou, Yuyang and Zhu, Y. X. and Huang, Yanping and Li, Yaohui and Zheng, Yi and Zhu, Yuchen and Ma, Yunxian and Tang, Ying and Zha, Yukun and Yan, Yuting and Ren, Z. Z. and Ren, Zehui and Sha, Zhangli and Fu, Zhe and Xu, Zhean and Xie, Zhenda and Zhang, Zhengyan and Hao, Zhewen and Ma, Zhicheng and Yan, Zhigang and Wu, Zhiyu and Gu, Zihui and Zhu, Zijia and Liu, Zijun and Li, Zilin and Xie, Ziwei and Song, Ziyang and Pan, Zizheng and Huang, Zhen and Xu, Zhipeng and Zhang, Zhongyu and Zhang, Zhen},
   year={2025},
   month=sep, pages={633–638} }

@inproceedings{kojima2022large,
title={Large Language Models are Zero-Shot Reasoners},
author={Takeshi Kojima and Shixiang Shane Gu and Machel Reid and Yutaka Matsuo and Yusuke Iwasawa},
booktitle={Advances in Neural Information Processing Systems},
editor={Alice H. Oh and Alekh Agarwal and Danielle Belgrave and Kyunghyun Cho},
year={2022},
url={https://openreview.net/forum?id=e2TBb5y0yFf}
}

@misc{wei2022finetunedlanguagemodelszeroshot,
      title={Finetuned Language Models Are Zero-Shot Learners}, 
      author={Jason Wei and Maarten Bosma and Vincent Y. Zhao and Kelvin Guu and Adams Wei Yu and Brian Lester and Nan Du and Andrew M. Dai and Quoc V. Le},
      year={2022},
      eprint={2109.01652},
      archivePrefix={arXiv},
      primaryClass={cs.CL},
      url={https://arxiv.org/abs/2109.01652}, 
}

@misc{chung2022scalinginstructionfinetunedlanguagemodels,
      title={Scaling Instruction-Finetuned Language Models}, 
      author={Hyung Won Chung and Le Hou and Shayne Longpre and Barret Zoph and Yi Tay and William Fedus and Yunxuan Li and Xuezhi Wang and Mostafa Dehghani and Siddhartha Brahma and Albert Webson and Shixiang Shane Gu and Zhuyun Dai and Mirac Suzgun and Xinyun Chen and Aakanksha Chowdhery and Alex Castro-Ros and Marie Pellat and Kevin Robinson and Dasha Valter and Sharan Narang and Gaurav Mishra and Adams Yu and Vincent Zhao and Yanping Huang and Andrew Dai and Hongkun Yu and Slav Petrov and Ed H. Chi and Jeff Dean and Jacob Devlin and Adam Roberts and Denny Zhou and Quoc V. Le and Jason Wei},
      year={2022},
      eprint={2210.11416},
      archivePrefix={arXiv},
      primaryClass={cs.LG},
      url={https://arxiv.org/abs/2210.11416}, 
}

@inproceedings{
sanh2022multitask,
title={Multitask Prompted Training Enables Zero-Shot Task Generalization},
author={Victor Sanh and Albert Webson and Colin Raffel and Stephen Bach and Lintang Sutawika and Zaid Alyafeai and Antoine Chaffin and Arnaud Stiegler and Arun Raja and Manan Dey and M Saiful Bari and Canwen Xu and Urmish Thakker and Shanya Sharma Sharma and Eliza Szczechla and Taewoon Kim and Gunjan Chhablani and Nihal Nayak and Debajyoti Datta and Jonathan Chang and Mike Tian-Jian Jiang and Han Wang and Matteo Manica and Sheng Shen and Zheng Xin Yong and Harshit Pandey and Rachel Bawden and Thomas Wang and Trishala Neeraj and Jos Rozen and Abheesht Sharma and Andrea Santilli and Thibault Fevry and Jason Alan Fries and Ryan Teehan and Teven Le Scao and Stella Biderman and Leo Gao and Thomas Wolf and Alexander M Rush},
booktitle={International Conference on Learning Representations},
year={2022},
url={https://openreview.net/forum?id=9Vrb9D0WI4}
}

@inproceedings{wang-etal-2022-super,
    title = "Super-{N}atural{I}nstructions: Generalization via Declarative Instructions on 1600+ {NLP} Tasks",
    author = "Wang, Yizhong  and
      Mishra, Swaroop  and
      Alipoormolabashi, Pegah  and
      Kordi, Yeganeh  and
      Mirzaei, Amirreza  and
      Naik, Atharva  and
      Ashok, Arjun  and
      Dhanasekaran, Arut Selvan  and
      Arunkumar, Anjana  and
      Stap, David  and
      Pathak, Eshaan  and
      Karamanolakis, Giannis  and
      Lai, Haizhi  and
      Purohit, Ishan  and
      Mondal, Ishani  and
      Anderson, Jacob  and
      Kuznia, Kirby  and
      Doshi, Krima  and
      Pal, Kuntal Kumar  and
      Patel, Maitreya  and
      Moradshahi, Mehrad  and
      Parmar, Mihir  and
      Purohit, Mirali  and
      Varshney, Neeraj  and
      Kaza, Phani Rohitha  and
      Verma, Pulkit  and
      Puri, Ravsehaj Singh  and
      Karia, Rushang  and
      Doshi, Savan  and
      Sampat, Shailaja Keyur  and
      Mishra, Siddhartha  and
      Reddy A, Sujan  and
      Patro, Sumanta  and
      Dixit, Tanay  and
      Shen, Xudong",
    editor = "Goldberg, Yoav  and
      Kozareva, Zornitsa  and
      Zhang, Yue",
    booktitle = "Proceedings of the 2022 Conference on Empirical Methods in Natural Language Processing",
    month = dec,
    year = "2022",
    address = "Abu Dhabi, United Arab Emirates",
    publisher = "Association for Computational Linguistics",
    url = "https://aclanthology.org/2022.emnlp-main.340/",
    doi = "10.18653/v1/2022.emnlp-main.340",
    pages = "5085--5109"
}

@inproceedings{
lin2025evaluation,
title={Evaluation and Benchmarking Suite for Financial Large Language Models and Agents},
author={Shengyuan Lin and Jaisal Patel and Qinchuan Zhang and Kaiwen He and Keyi Wang and Yan Wang and Matt White and Kairong Xiao and Xiao-Yang Liu},
booktitle={NeurIPS 2025 Workshop on Evaluating the Evolving LLM Lifecycle: Benchmarks, Emergent Abilities, and Scaling},
year={2025},
url={https://openreview.net/forum?id=sSY4h3MFUB}
}

@article {Gaber2024.09.27.24314505,
	author = {Gaber, Farieda and Shaik, Maqsood and Franke, Vedran and Akalin, Altuna},
	title = {Evaluating large language model workflows in clinical decision support: referral, triage, and diagnosis},
	elocation-id = {2024.09.27.24314505},
	year = {2024},
	doi = {10.1101/2024.09.27.24314505},
	publisher = {Cold Spring Harbor Laboratory Press},
	URL = {https://www.medrxiv.org/content/early/2024/09/28/2024.09.27.24314505},
	eprint = {https://www.medrxiv.org/content/early/2024/09/28/2024.09.27.24314505.full.pdf},
	journal = {medRxiv}
}

@misc{jiang2025investigatinglargelanguagemodels,
      title={Investigating Large Language Models for Code Vulnerability Detection: An Experimental Study}, 
      author={Xuefeng Jiang and Lvhua Wu and Sheng Sun and Jia Li and Jingjing Xue and Yuwei Wang and Tingting Wu and Min Liu},
      year={2025},
      eprint={2412.18260},
      archivePrefix={arXiv},
      primaryClass={cs.CL},
      url={https://arxiv.org/abs/2412.18260}, 
}

@inproceedings{
guha2023legalbench,
title={LegalBench: A Collaboratively Built Benchmark for Measuring Legal Reasoning in Large Language Models},
author={Neel Guha and Julian Nyarko and Daniel E. Ho and Christopher Re and Adam Chilton and Aditya Narayana and Alex Chohlas-Wood and Austin Peters and Brandon Waldon and Daniel Rockmore and Diego Zambrano and Dmitry Talisman and Enam Hoque and Faiz Surani and Frank Fagan and Galit Sarfaty and Gregory M. Dickinson and Haggai Porat and Jason Hegland and Jessica Wu and Joe Nudell and Joel Niklaus and John J Nay and Jonathan H. Choi and Kevin Tobia and Margaret Hagan and Megan Ma and Michael Livermore and Nikon Rasumov-Rahe and Nils Holzenberger and Noam Kolt and Peter Henderson and Sean Rehaag and Sharad Goel and Shang Gao and Spencer Williams and Sunny Gandhi and Tom Zur and Varun Iyer and Zehua Li},
booktitle={Thirty-seventh Conference on Neural Information Processing Systems Datasets and Benchmarks Track},
year={2023},
url={https://openreview.net/forum?id=WqSPQFxFRC}
}

@misc{kant2025robustlegalreasoningharnessing,
      title={Towards Robust Legal Reasoning: Harnessing Logical LLMs in Law}, 
      author={Manuj Kant and Sareh Nabi and Manav Kant and Roland Scharrer and Megan Ma and Marzieh Nabi},
      year={2025},
      eprint={2502.17638},
      archivePrefix={arXiv},
      primaryClass={cs.CY},
      url={https://arxiv.org/abs/2502.17638}, 
}

@inproceedings{NEURIPS2024_70702e8c,
 author = {Mehrotra, Anay and Zampetakis, Manolis and Kassianik, Paul and Nelson, Blaine and Anderson, Hyrum and Singer, Yaron and Karbasi, Amin},
 booktitle = {Advances in Neural Information Processing Systems},
 doi = {10.52202/079017-1952},
 editor = {A. Globerson and L. Mackey and D. Belgrave and A. Fan and U. Paquet and J. Tomczak and C. Zhang},
 pages = {61065--61105},
 publisher = {Curran Associates, Inc.},
 title = {Tree of Attacks: Jailbreaking Black-Box LLMs Automatically},
 url = {https://proceedings.neurips.cc/paper_files/paper/2024/file/70702e8cbb4890b4a467b984ae59828a-Paper-Conference.pdf},
 volume = {37},
 year = {2024}
}

@inproceedings{
wei2023jailbroken,
title={Jailbroken: How Does {LLM} Safety Training Fail?},
author={Alexander Wei and Nika Haghtalab and Jacob Steinhardt},
booktitle={Thirty-seventh Conference on Neural Information Processing Systems},
year={2023},
url={https://openreview.net/forum?id=jA235JGM09}
}

@misc{heretic,
  author = {Weidmann, Philipp Emanuel},
  title = {Heretic: Fully automatic censorship removal for language models},
  year = {2025},
  publisher = {GitHub},
  journal = {GitHub repository},
  howpublished = {\url{https://github.com/p-e-w/heretic}}
}

@misc{lai2025projected_abliteration,
  author       = {Jim Lai},
  title        = {Projected Abliteration},
  year         = {2025},
  month        = {October},
  howpublished = {\url{https://huggingface.co/blog/grimjim/projected-abliteration}},
  note         = {Hugging Face Blog (Community Article)}
}

@misc{eleutherai_diff_in_means,
  author       = {{EleutherAI}},
  title        = {Diff-in-Means},
  year         = {2023},
  howpublished = {\url{https://blog.eleuther.ai/diff-in-means/}},
  note         = {Accessed: 2026-04-16}
}

@misc{vaswani2023attentionneed,
      title={Attention Is All You Need}, 
      author={Ashish Vaswani and Noam Shazeer and Niki Parmar and Jakob Uszkoreit and Llion Jones and Aidan N. Gomez and Lukasz Kaiser and Illia Polosukhin},
      year={2023},
      eprint={1706.03762},
      archivePrefix={arXiv},
      primaryClass={cs.CL},
      url={https://arxiv.org/abs/1706.03762}, 
}

@inproceedings{
j.2018generating,
title={Generating Wikipedia by Summarizing Long Sequences},
author={Peter J. Liu* and Mohammad Saleh* and Etienne Pot and Ben Goodrich and Ryan Sepassi and Lukasz Kaiser and Noam Shazeer},
booktitle={International Conference on Learning Representations},
year={2018},
url={https://openreview.net/forum?id=Hyg0vbWC-},
}

@article{elhage2021mathematical,
   title={A Mathematical Framework for Transformer Circuits},
   author={Elhage, Nelson and Nanda, Neel and Olsson, Catherine and Henighan, Tom and Joseph, Nicholas and Mann, Ben and Askell, Amanda and Bai, Yuntao and Chen, Anna and Conerly, Tom and DasSarma, Nova and Drain, Dawn and Ganguli, Deep and Hatfield-Dodds, Zac and Hernandez, Danny and Jones, Andy and Kernion, Jackson and Lovitt, Liane and Ndousse, Kamal and Amodei, Dario and Brown, Tom and Clark, Jack and Kaplan, Jared and McCandlish, Sam and Olah, Chris},
   year={2021},
   journal={Transformer Circuits Thread},
   note={https://transformer-circuits.pub/2021/framework/index.html}
}

@inproceedings{rimsky-etal-2024-steering,
    title = "Steering Llama 2 via Contrastive Activation Addition",
    author = "Rimsky, Nina  and
      Gabrieli, Nick  and
      Schulz, Julian  and
      Tong, Meg  and
      Hubinger, Evan  and
      Turner, Alexander",
    editor = "Ku, Lun-Wei  and
      Martins, Andre  and
      Srikumar, Vivek",
    booktitle = "Proceedings of the 62nd Annual Meeting of the Association for Computational Linguistics (Volume 1: Long Papers)",
    month = aug,
    year = "2024",
    address = "Bangkok, Thailand",
    publisher = "Association for Computational Linguistics",
    url = "https://aclanthology.org/2024.acl-long.828/",
    doi = "10.18653/v1/2024.acl-long.828",
    pages = "15504--15522"
}

@inproceedings{
postmus2024steering,
title={Steering Large Language Models using Conceptors: Improving Addition-Based Activation Engineering},
author={Joris Postmus and Steven Abreu},
booktitle = {MINT: Foundation Model Interventions},
year={2024},
url={https://openreview.net/forum?id=gyAnAq16HC}
}

@misc{zou2025representationengineeringtopdownapproach,
      title={Representation Engineering: A Top-Down Approach to AI Transparency}, 
      author={Andy Zou and Long Phan and Sarah Chen and James Campbell and Phillip Guo and Richard Ren and Alexander Pan and Xuwang Yin and Mantas Mazeika and Ann-Kathrin Dombrowski and Shashwat Goel and Nathaniel Li and Michael J. Byun and Zifan Wang and Alex Mallen and Steven Basart and Sanmi Koyejo and Dawn Song and Matt Fredrikson and J. Zico Kolter and Dan Hendrycks},
      year={2025},
      eprint={2310.01405},
      archivePrefix={arXiv},
      primaryClass={cs.LG},
      url={https://arxiv.org/abs/2310.01405}, 
}

@inproceedings{
zou2024improving,
title={Improving Alignment and Robustness with Circuit Breakers},
author={Andy Zou and Long Phan and Justin Wang and Derek Duenas and Maxwell Lin and Maksym Andriushchenko and J Zico Kolter and Matt Fredrikson and Dan Hendrycks},
booktitle={The Thirty-eighth Annual Conference on Neural Information Processing Systems},
year={2024},
url={https://openreview.net/forum?id=IbIB8SBKFV}
}

@inproceedings{
lee2025programming,
title={Programming Refusal with Conditional Activation Steering},
author={Bruce W. Lee and Inkit Padhi and Karthikeyan Natesan Ramamurthy and Erik Miehling and Pierre Dognin and Manish Nagireddy and Amit Dhurandhar},
booktitle={The Thirteenth International Conference on Learning Representations},
year={2025},
url={https://openreview.net/forum?id=Oi47wc10sm}
}

@inproceedings{
shen2025jailbreak,
title={Jailbreak Antidote: Runtime Safety-Utility Balance via Sparse Representation Adjustment in Large Language Models},
author={Guobin Shen and Dongcheng Zhao and Yiting Dong and Xiang He and Yi Zeng},
booktitle={The Thirteenth International Conference on Learning Representations},
year={2025},
url={https://openreview.net/forum?id=s20W12XTF8}
}

@misc{wang2025surgicalcheapflexiblemitigating,
      title={Surgical, Cheap, and Flexible: Mitigating False Refusal in Language Models via Single Vector Ablation}, 
      author={Xinpeng Wang and Chengzhi Hu and Paul Röttger and Barbara Plank},
      year={2025},
      eprint={2410.03415},
      archivePrefix={arXiv},
      primaryClass={cs.CL},
      url={https://arxiv.org/abs/2410.03415}, 
}

@inproceedings{
sheng2026alphasteer,
title={AlphaSteer: Learning Refusal Steering with Principled Null-Space Constraint},
author={Leheng Sheng and Changshuo Shen and Weixiang Zhao and Junfeng Fang and Xiaohao Liu and Zhenkai Liang and Xiang Wang and An Zhang and Tat-Seng Chua},
booktitle={The Fourteenth International Conference on Learning Representations},
year={2026},
url={https://openreview.net/forum?id=1vvbzAqdTe}
}

@misc{
shairah2026turning,
title={Turning the Spell Around: Lightweight Alignment Amplification via Rank-One Safety Injection},
author={Harethah Abu Shairah and Hasan Abed Al Kader Hammoud and George Turkiyyah and Bernard Ghanem},
year={2026},
url={https://openreview.net/forum?id=8c2SbG5PLj}
}

@misc{openai2024gpt4technicalreport,
      title={GPT-4 Technical Report}, 
      author={OpenAI and Josh Achiam and Steven Adler and Sandhini Agarwal and Lama Ahmad and Ilge Akkaya and Florencia Leoni Aleman and Diogo Almeida and Janko Altenschmidt and Sam Altman and Shyamal Anadkat and Red Avila and Igor Babuschkin and Suchir Balaji and Valerie Balcom and Paul Baltescu and Haiming Bao and Mohammad Bavarian and Jeff Belgum and Irwan Bello and Jake Berdine and Gabriel Bernadett-Shapiro and Christopher Berner and Lenny Bogdonoff and Oleg Boiko and Madelaine Boyd and Anna-Luisa Brakman and Greg Brockman and Tim Brooks and Miles Brundage and Kevin Button and Trevor Cai and Rosie Campbell and Andrew Cann and Brittany Carey and Chelsea Carlson and Rory Carmichael and Brooke Chan and Che Chang and Fotis Chantzis and Derek Chen and Sully Chen and Ruby Chen and Jason Chen and Mark Chen and Ben Chess and Chester Cho and Casey Chu and Hyung Won Chung and Dave Cummings and Jeremiah Currier and Yunxing Dai and Cory Decareaux and Thomas Degry and Noah Deutsch and Damien Deville and Arka Dhar and David Dohan and Steve Dowling and Sheila Dunning and Adrien Ecoffet and Atty Eleti and Tyna Eloundou and David Farhi and Liam Fedus and Niko Felix and Simón Posada Fishman and Juston Forte and Isabella Fulford and Leo Gao and Elie Georges and Christian Gibson and Vik Goel and Tarun Gogineni and Gabriel Goh and Rapha Gontijo-Lopes and Jonathan Gordon and Morgan Grafstein and Scott Gray and Ryan Greene and Joshua Gross and Shixiang Shane Gu and Yufei Guo and Chris Hallacy and Jesse Han and Jeff Harris and Yuchen He and Mike Heaton and Johannes Heidecke and Chris Hesse and Alan Hickey and Wade Hickey and Peter Hoeschele and Brandon Houghton and Kenny Hsu and Shengli Hu and Xin Hu and Joost Huizinga and Shantanu Jain and Shawn Jain and Joanne Jang and Angela Jiang and Roger Jiang and Haozhun Jin and Denny Jin and Shino Jomoto and Billie Jonn and Heewoo Jun and Tomer Kaftan and Łukasz Kaiser and Ali Kamali and Ingmar Kanitscheider and Nitish Shirish Keskar and Tabarak Khan and Logan Kilpatrick and Jong Wook Kim and Christina Kim and Yongjik Kim and Jan Hendrik Kirchner and Jamie Kiros and Matt Knight and Daniel Kokotajlo and Łukasz Kondraciuk and Andrew Kondrich and Aris Konstantinidis and Kyle Kosic and Gretchen Krueger and Vishal Kuo and Michael Lampe and Ikai Lan and Teddy Lee and Jan Leike and Jade Leung and Daniel Levy and Chak Ming Li and Rachel Lim and Molly Lin and Stephanie Lin and Mateusz Litwin and Theresa Lopez and Ryan Lowe and Patricia Lue and Anna Makanju and Kim Malfacini and Sam Manning and Todor Markov and Yaniv Markovski and Bianca Martin and Katie Mayer and Andrew Mayne and Bob McGrew and Scott Mayer McKinney and Christine McLeavey and Paul McMillan and Jake McNeil and David Medina and Aalok Mehta and Jacob Menick and Luke Metz and Andrey Mishchenko and Pamela Mishkin and Vinnie Monaco and Evan Morikawa and Daniel Mossing and Tong Mu and Mira Murati and Oleg Murk and David Mély and Ashvin Nair and Reiichiro Nakano and Rajeev Nayak and Arvind Neelakantan and Richard Ngo and Hyeonwoo Noh and Long Ouyang and Cullen O'Keefe and Jakub Pachocki and Alex Paino and Joe Palermo and Ashley Pantuliano and Giambattista Parascandolo and Joel Parish and Emy Parparita and Alex Passos and Mikhail Pavlov and Andrew Peng and Adam Perelman and Filipe de Avila Belbute Peres and Michael Petrov and Henrique Ponde de Oliveira Pinto and Michael and Pokorny and Michelle Pokrass and Vitchyr H. Pong and Tolly Powell and Alethea Power and Boris Power and Elizabeth Proehl and Raul Puri and Alec Radford and Jack Rae and Aditya Ramesh and Cameron Raymond and Francis Real and Kendra Rimbach and Carl Ross and Bob Rotsted and Henri Roussez and Nick Ryder and Mario Saltarelli and Ted Sanders and Shibani Santurkar and Girish Sastry and Heather Schmidt and David Schnurr and John Schulman and Daniel Selsam and Kyla Sheppard and Toki Sherbakov and Jessica Shieh and Sarah Shoker and Pranav Shyam and Szymon Sidor and Eric Sigler and Maddie Simens and Jordan Sitkin and Katarina Slama and Ian Sohl and Benjamin Sokolowsky and Yang Song and Natalie Staudacher and Felipe Petroski Such and Natalie Summers and Ilya Sutskever and Jie Tang and Nikolas Tezak and Madeleine B. Thompson and Phil Tillet and Amin Tootoonchian and Elizabeth Tseng and Preston Tuggle and Nick Turley and Jerry Tworek and Juan Felipe Cerón Uribe and Andrea Vallone and Arun Vijayvergiya and Chelsea Voss and Carroll Wainwright and Justin Jay Wang and Alvin Wang and Ben Wang and Jonathan Ward and Jason Wei and CJ Weinmann and Akila Welihinda and Peter Welinder and Jiayi Weng and Lilian Weng and Matt Wiethoff and Dave Willner and Clemens Winter and Samuel Wolrich and Hannah Wong and Lauren Workman and Sherwin Wu and Jeff Wu and Michael Wu and Kai Xiao and Tao Xu and Sarah Yoo and Kevin Yu and Qiming Yuan and Wojciech Zaremba and Rowan Zellers and Chong Zhang and Marvin Zhang and Shengjia Zhao and Tianhao Zheng and Juntang Zhuang and William Zhuk and Barret Zoph},
      year={2024},
      eprint={2303.08774},
      archivePrefix={arXiv},
      primaryClass={cs.CL},
      url={https://arxiv.org/abs/2303.08774}, 
}

@inproceedings{ouyang2022training,
title={Training language models to follow instructions with human feedback},
author={Long Ouyang and Jeffrey Wu and Xu Jiang and Diogo Almeida and Carroll Wainwright and Pamela Mishkin and Chong Zhang and Sandhini Agarwal and Katarina Slama and Alex Gray and John Schulman and Jacob Hilton and Fraser Kelton and Luke Miller and Maddie Simens and Amanda Askell and Peter Welinder and Paul Christiano and Jan Leike and Ryan Lowe},
booktitle={Advances in Neural Information Processing Systems},
editor={Alice H. Oh and Alekh Agarwal and Danielle Belgrave and Kyunghyun Cho},
year={2022},
url={https://openreview.net/forum?id=TG8KACxEON}
}

@inproceedings{
bianchi2024safetytuned,
title={Safety-Tuned {LL}a{MA}s: Lessons From Improving the Safety of Large Language Models that Follow Instructions},
author={Federico Bianchi and Mirac Suzgun and Giuseppe Attanasio and Paul Rottger and Dan Jurafsky and Tatsunori Hashimoto and James Zou},
booktitle={The Twelfth International Conference on Learning Representations},
year={2024},
url={https://openreview.net/forum?id=gT5hALch9z}
}

@misc{bai2022constitutionalaiharmlessnessai,
      title={Constitutional AI: Harmlessness from AI Feedback}, 
      author={Yuntao Bai and Saurav Kadavath and Sandipan Kundu and Amanda Askell and Jackson Kernion and Andy Jones and Anna Chen and Anna Goldie and Azalia Mirhoseini and Cameron McKinnon and Carol Chen and Catherine Olsson and Christopher Olah and Danny Hernandez and Dawn Drain and Deep Ganguli and Dustin Li and Eli Tran-Johnson and Ethan Perez and Jamie Kerr and Jared Mueller and Jeffrey Ladish and Joshua Landau and Kamal Ndousse and Kamile Lukosuite and Liane Lovitt and Michael Sellitto and Nelson Elhage and Nicholas Schiefer and Noemi Mercado and Nova DasSarma and Robert Lasenby and Robin Larson and Sam Ringer and Scott Johnston and Shauna Kravec and Sheer El Showk and Stanislav Fort and Tamera Lanham and Timothy Telleen-Lawton and Tom Conerly and Tom Henighan and Tristan Hume and Samuel R. Bowman and Zac Hatfield-Dodds and Ben Mann and Dario Amodei and Nicholas Joseph and Sam McCandlish and Tom Brown and Jared Kaplan},
      year={2022},
      eprint={2212.08073},
      archivePrefix={arXiv},
      primaryClass={cs.CL},
      url={https://arxiv.org/abs/2212.08073}, 
}

@inproceedings{
liu2024autodan,
title={Auto{DAN}: Generating Stealthy Jailbreak Prompts on Aligned Large Language Models},
author={Xiaogeng Liu and Nan Xu and Muhao Chen and Chaowei Xiao},
booktitle={The Twelfth International Conference on Learning Representations},
year={2024},
url={https://openreview.net/forum?id=7Jwpw4qKkb}
}

@inproceedings{
chao2023jailbreaking,
title={Jailbreaking Black Box Large Language Models in Twenty Queries},
author={Patrick Chao and Alexander Robey and Edgar Dobriban and Hamed Hassani and George J. Pappas and Eric Wong},
booktitle={R0-FoMo:Robustness of Few-shot and Zero-shot Learning in Large Foundation Models},
year={2023},
url={https://openreview.net/forum?id=rYWD5TMaLj}
}

@misc{zou2023universaltransferableadversarialattacks,
      title={Universal and Transferable Adversarial Attacks on Aligned Language Models}, 
      author={Andy Zou and Zifan Wang and Nicholas Carlini and Milad Nasr and J. Zico Kolter and Matt Fredrikson},
      year={2023},
      eprint={2307.15043},
      archivePrefix={arXiv},
      primaryClass={cs.CL},
      url={https://arxiv.org/abs/2307.15043}, 
}

@inproceedings{NIPS2017_d5e2c0ad,
 author = {Christiano, Paul F and Leike, Jan and Brown, Tom and Martic, Miljan and Legg, Shane and Amodei, Dario},
 booktitle = {Advances in Neural Information Processing Systems},
 editor = {I. Guyon and U. Von Luxburg and S. Bengio and H. Wallach and R. Fergus and S. Vishwanathan and R. Garnett},
 pages = {},
 publisher = {Curran Associates, Inc.},
 title = {Deep Reinforcement Learning from Human Preferences},
 url = {https://proceedings.neurips.cc/paper_files/paper/2017/file/d5e2c0adad503c91f91df240d0cd4e49-Paper.pdf},
 volume = {30},
 year = {2017}
}

@inproceedings{
rafailov2023direct,
title={Direct Preference Optimization: Your Language Model is Secretly a Reward Model},
author={Rafael Rafailov and Archit Sharma and Eric Mitchell and Christopher D Manning and Stefano Ermon and Chelsea Finn},
booktitle={Thirty-seventh Conference on Neural Information Processing Systems},
year={2023},
url={https://openreview.net/forum?id=HPuSIXJaa9}
}

@inproceedings{
park2023the,
title={The Linear Representation Hypothesis and the Geometry of Large Language Models},
author={Kiho Park and Yo Joong Choe and Victor Veitch},
booktitle={Causal Representation Learning Workshop at NeurIPS 2023},
year={2023},
url={https://openreview.net/forum?id=T0PoOJg8cK}
}

@online{durmus2024steering,
author = {Esin Durmus and Alex Tamkin and Jack Clark and Jerry Wei and Jonathan Marcus and Joshua Batson and Kunal Handa and Liane Lovitt and Meg Tong and Miles McCain and Oliver Rausch and Saffron Huang and Sam Bowman and Stuart Ritchie and Tom Henighan and Deep Ganguli},
title = {Evaluating Feature Steering: A Case Study in Mitigating Social Biases},
date = {2024-10-25},
year = {2024},
url = {https://anthropic.com/research/evaluating-feature-steering},
}

@article{elhage2022superposition,
   title={Toy Models of Superposition},
   author={Elhage, Nelson and Hume, Tristan and Olsson, Catherine and Schiefer, Nicholas and Henighan, Tom and Kravec, Shauna and Hatfield-Dodds, Zac and Lasenby, Robert and Drain, Dawn and Chen, Carol and Grosse, Roger and McCandlish, Sam and Kaplan, Jared and Amodei, Dario and Wattenberg, Martin and Olah, Christopher},
   year={2022},
   journal={Transformer Circuits Thread},
   url={https://transformer-circuits.pub/2022/toy_model/index.html}
}

@misc{obrien2025steeringlanguagemodelrefusal,
      title={Steering Language Model Refusal with Sparse Autoencoders}, 
      author={Kyle O'Brien and David Majercak and Xavier Fernandes and Richard Edgar and Blake Bullwinkel and Jingya Chen and Harsha Nori and Dean Carignan and Eric Horvitz and Forough Poursabzi-Sangdeh},
      year={2025},
      eprint={2411.11296},
      archivePrefix={arXiv},
      primaryClass={cs.LG},
      url={https://arxiv.org/abs/2411.11296}, 
}

@inproceedings{
meng2022locating,
title={Locating and Editing Factual Associations in {GPT}},
author={Kevin Meng and David Bau and Alex J Andonian and Yonatan Belinkov},
booktitle={Advances in Neural Information Processing Systems},
editor={Alice H. Oh and Alekh Agarwal and Danielle Belgrave and Kyunghyun Cho},
year={2022},
url={https://openreview.net/forum?id=-h6WAS6eE4}
}

@inproceedings{
meng2023massediting,
title={Mass-Editing Memory in a Transformer},
author={Kevin Meng and Arnab Sen Sharma and Alex J Andonian and Yonatan Belinkov and David Bau},
booktitle={The Eleventh International Conference on Learning Representations },
year={2023},
url={https://openreview.net/forum?id=MkbcAHIYgyS}
}

@inproceedings{
ilharco2023editing,
title={Editing models with task arithmetic},
author={Gabriel Ilharco and Marco Tulio Ribeiro and Mitchell Wortsman and Ludwig Schmidt and Hannaneh Hajishirzi and Ali Farhadi},
booktitle={The Eleventh International Conference on Learning Representations },
year={2023},
url={https://openreview.net/forum?id=6t0Kwf8-jrj}
}

@misc{shi2024largelanguagemodelsafety,
      title={Large Language Model Safety: A Holistic Survey}, 
      author={Dan Shi and Tianhao Shen and Yufei Huang and Zhigen Li and Yongqi Leng and Renren Jin and Chuang Liu and Xinwei Wu and Zishan Guo and Linhao Yu and Ling Shi and Bojian Jiang and Deyi Xiong},
      year={2024},
      eprint={2412.17686},
      archivePrefix={arXiv},
      primaryClass={cs.AI},
      url={https://arxiv.org/abs/2412.17686}, 
}

@misc{mazeika2024harmbenchstandardizedevaluationframework,
      title={HarmBench: A Standardized Evaluation Framework for Automated Red Teaming and Robust Refusal}, 
      author={Mantas Mazeika and Long Phan and Xuwang Yin and Andy Zou and Zifan Wang and Norman Mu and Elham Sakhaee and Nathaniel Li and Steven Basart and Bo Li and David Forsyth and Dan Hendrycks},
      year={2024},
      eprint={2402.04249},
      archivePrefix={arXiv},
      primaryClass={cs.LG},
      url={https://arxiv.org/abs/2402.04249}, 
}

\medskip









\end{document}